\documentclass{article} 
\usepackage{iclr2024_conference,times}

\usepackage{url}
\usepackage{booktabs}
\usepackage{colortbl}
\usepackage{color,bm}
\usepackage{algorithm,amsmath,amssymb,bm,amsthm}
\usepackage{multirow}

\usepackage{algorithmic}
\usepackage{graphicx}
\usepackage{wrapfig,lipsum,booktabs}
\usepackage{amsmath,amssymb,amsfonts}
\usepackage{enumitem}
\usepackage{xcolor}
\definecolor{linkcolor}{RGB}{0, 0, 0}
\usepackage[colorlinks=true,allcolors=linkcolor,pageanchor=true,plainpages=false,pdfpagelabels,bookmarks,bookmarksnumbered]{hyperref}
\usepackage{wrapfig,lipsum,booktabs}
\title{Entity-Aware Biaffine Attention Model for Improved Constituent Parsing with Reduced Entity Violations}

\author{Xinyi Bai\\
 Henan Institute of Technology, China 
}

\iclrfinalcopy 
\begin{document}

\maketitle

\begin{abstract}
Constituency parsing involves analyzing a sentence by breaking it into sub-phrases, or constituents. While many deep neural models have achieved state-of-the-art performance in this task, they often overlook the entity-violating issue, where an entity fails to form a complete sub-tree in the resultant parsing tree. To address this, we propose an entity-aware biaffine attention model for constituent parsing. This model incorporates entity information into the biaffine attention mechanism by using additional entity role vectors for potential phrases, which enhances the parsing accuracy. We introduce a new metric, the Entity Violating Rate (EVR), to quantify the extent of entity violations in parsing results. Experiments on three popular datasets—ONTONOTES, PTB, and CTB—demonstrate that our model achieves the lowest EVR while maintaining high precision, recall, and F1-scores comparable to existing models. Further evaluation in downstream tasks, such as sentence sentiment analysis, highlights the effectiveness of our model and the validity of the proposed EVR metric.

\end{abstract}

\section{Introduction}
\begin{figure}
\centering
\setlength{\abovecaptionskip}{-0.2cm}
\includegraphics[width=1.0 \textwidth]{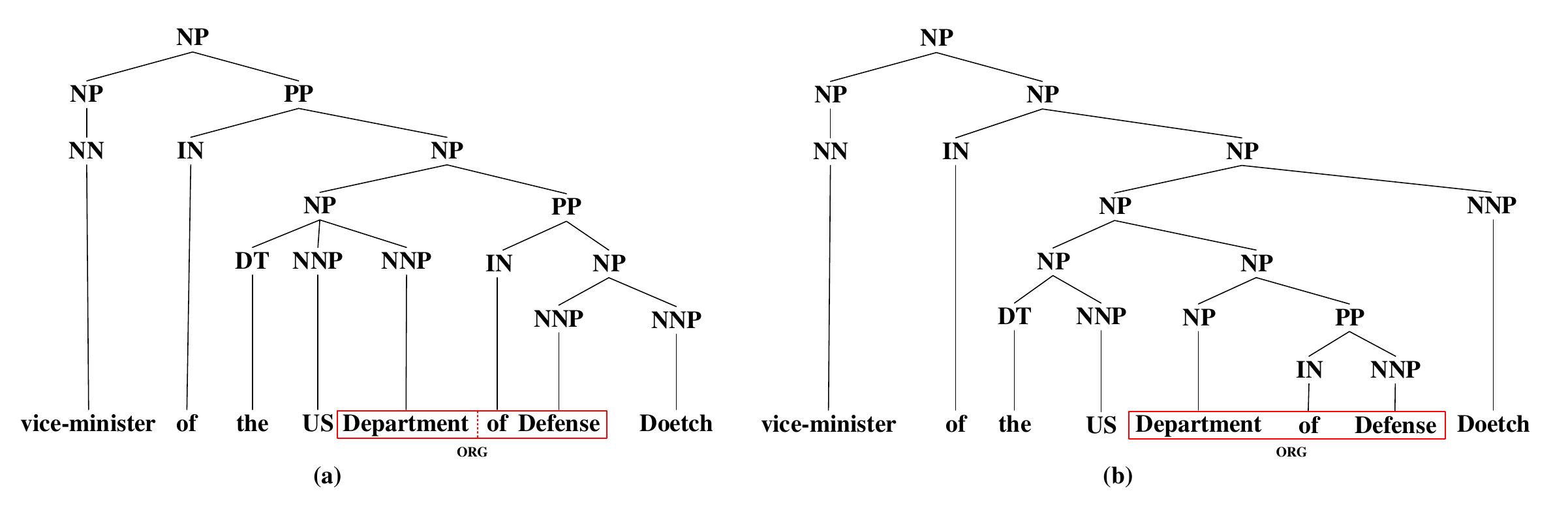}
\caption{Constituent parsing tree of phrase "vice-minister of the US Department of Defense Doetch". (a) The entity-violating case. (b) The entity consistent case. } \label{fig1}
\end{figure}

Constituent parsing is to construct the syntactic tree for a given sentence whose words constitute leaf nodes. In the syntactic tree, non-terminal nodes are called constituents. For intuition, Fig.~\ref{fig1}  (a) illustrates a constituent parsing tree of the phrase “vice-minister of the US Department of Defense Doetch”. This tree serves as an important feature to represent a sentence, and has been applied in many high-level natural language tasks, such as sentiment analysis \cite{kim2019dynamic}, relation extraction \cite{jiang2019constituency}, natural language inference \cite{chen2016enhanced}, and machine translation \cite{ma2018forest}.

In this subject recent chart-based neural models have achieved state-of-the-art results by using advanced text encoders (i.e., BERT and XLNet) to represent all possible spans, where each span stands for a single word or several consecutive words from a sentence. Such models score each span, and then employs CKY algorithm to choose the resultant tree in terms of the highest score \cite{kitaev2018multilingual,kitaev2018constituency,mrini2019rethinking,tian2020improving}.
Despite their astonishing success in light of precision, recall and F1-score, few consider whether such arts also work well in the entity-violating issue: the entity span parsed by a neural model does not conform to the true phrasal one of human natural language. As in Fig. \ref{fig1}, “Department of Defense” as a true organization (ORG) entity should be in a complete sub-tree (b) but is parsed by previous neural models so that its words are distributed into two separate sub-trees (a). 

Finkel et al. \cite{finkel2009joint,finkel2010hierarchical} firstly focus on this issue. They manually annotate the original dataset ONTONOTES \cite{hovy2006ontonotes} by adding entity nodes to the constituent parsing trees. This new dataset can greatly promote the model to output the consistent entity spans. Besides, other works~\cite{li2013improved,li2018integrated} endeavor to explore the entity-related label as heuristic information, for instance, Chinese word `SHI' can be labeled with `GPE-END'. Different from aforesaid works that absorb entity-related information, Yang et al. \cite{wang2019chinese} considered utilizing the named entity recognition (NER) task to capture the entity information, and combining the PCFG algorithm for parsing, of which the parameter sharing enjoys the benefits of two tasks. However, since these entity-related parsing models either demand manual annotations or implicitly merge entity information into training process of constituent parsing, it may be underestimate the potential of the entity information. 

To attack such problems above, we propose a biaffine attention based parsing model by integrating entity information into constituent parsing process without any manual annotations. Our model overally follows the basic biaffine attention model \cite{zhang2020fast}. Similarly, each word is represented as two role vectors: start role vector and end role vector. Differently, we encode the entity information as the entity role vector and append it to the previous two role vectors. In this simple way, the proposed vector is more informative and reasonable than before. By using the proposed entity-aware vector for each word, the basic biaffine attention model can boost potential score for each entity span. In our empirical studies, we find that although the entity information reinforces the ability of attention model to parse consistent phrasal span, it is lack of effective supervision and the entity information might be ignored in the learning process. To this end, we treat a simple NER task as the supervision similar to \cite{yu2020named}. In our model, we add a binary NER model to share the same word embeddings with the primary parsing model. It helps our parsing model to capture informative entity structures. The overall model is light-weight, simple yet effective, free from large amounts of manual annotations, and even achieves the state-of-the-art performance on the benchmarks. To evaluate the entity-violating degree, we propose an very intuitive metric to calculate the ratio of the entity-violating spans to the entire samples.

In summary, this paper makes the following contributions.
\begin{itemize} 
  \item [1)] We put forward an entity-aware biaffine attention model for constituent parsing, which encodes the entity information of a span as the attention input component.
  \item [2)] To further exert the entity information in the attention model, we introduce an auxiliary bi-nary NER model in the whole parsing model, in order to make the parsing model aware of entity information.
  \item [3)] Experiments on three datasets including ONTONOTES, PTB and CTB show that our strategy greatly promote the parsing performance, especially based on entity-violating metric. More importantly, the proposed model achieves the sound performance in terms of precision, recall and F1-score. 
  \item [4)] To make our model more convincing, we apply our parsing model for a typical downstream task---sentence classification \cite{kim2019dynamic}. Extensive results verify that our model performs best when comparing with several well-behaved parsing siblings. 
\end{itemize}

\section{Related work}
\subsection{Constituent Parsing.}
Most constituent parsing models can be roughly classified into three types: transition-based, sequence-based and chart-based ones, according to which decoding algorithm they choose. For transition-based models, without no need of the decoder, they directly build the parsing trees through a sequence of ‘shift’ or ‘reduce’ actions, or other extended actions \cite{yin2024continuous}. A recent representative model proposed by Yang and Deng starts with current partial parsing tree, then builds the corresponding graph and applies dynamic graph neural network to decide the next action \cite{yang2020strongly,yin2024dynamic,ju2024survey}. This process will continuously iterate until the entire parsing tree is built successfully. Sequence-based approaches primarily convert a tree structure into a linear sequence form and generate a sequence of pertinent labels using seq2seq model \cite{zhang2022would,zhang2023investigating,meng2024deep,yin2023dream,vinyals2014grammar,strzyz2019sequence}. Differently, chart-based methods firstly score all possible spans, then use CKY dynamic algorithm to decode out a highest scored tree \cite{yin2022generic,ai2023gcn,shou2023adversarial,tang2024merging,kitaev2018constituency,mrini2019rethinking,tian2020improving,zhang2020fast}. In this regard, attention mechanism is introduced to represent a span by using self-attention \cite{kitaev2018constituency,yin2023omg,yin2023messages}, labeling attention \cite{mrini2019rethinking,pang2023sa,yin2023coco} and n-gram attention \cite{tian2020improving} to complement those span boundary information. These models focus on how to construct a more powerful word representation for model inference. 
However, none of these models above consider entity-violating issue. Although these models take entity information into consideration \cite{finkel2009joint,finkel2010hierarchical,li2013improved,li2018integrated,wang2019chinese}, they are either suffer from massive manual labeling work \cite{finkel2009joint,yin2022deal,li2013improved,li2018integrated} or cannot effectively convey entity information directly in parsing process \cite{wang2019chinese,yin2022dynamic,yin2022generic}. To this end, we introduce an entity-aware parsing model which can apply entity information directly into inference process without extra labeling work.

\section{Method}

\subsection{Parsing model}

This section explores the entity-aware biaffine attention model for constituent parsing, which follows the two-step parsing model \cite{zhang2020fast}. The first step (span-parsing) decides whether a span is a node in the resultant constituent tree, and the second label-parsing model labels each node with POS tag. The two steps share the same contextual word embeddings.

\begin{figure}
\setlength{\abovecaptionskip}{-0.1cm}
\centering
\includegraphics[width=0.8\textwidth]{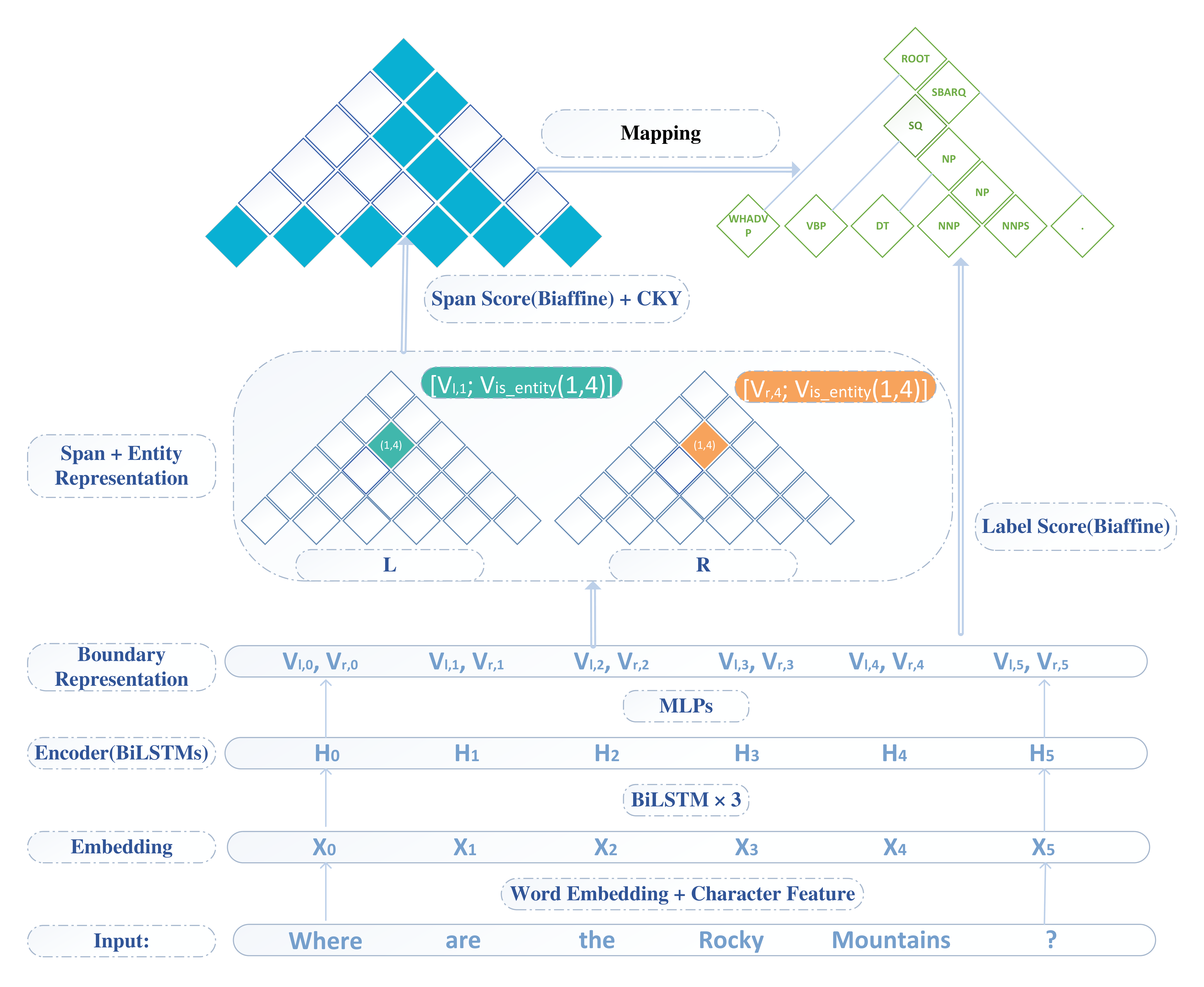}
\caption{The framework of entity-aware constituent parsing approach} \label{fig2}
\end{figure}

Figure~\ref{fig2} shows the overall architecture of constituent parsing. For the given sentence, the first input embedding for each word (e.g. $X_i$) is the concatenation of word embedding, and its char-level feature gained through CharLSTM on each word. To obtain contextual information, we feed the embedding of each word into a 3-layer BiLSTM layer. The BiLSTM module outputs two hidden embeddings $f$ and $b$ for each word from forward and backward direction. Then for each word, e.g. $word_i$ we concatenate $f_i$ and $b_{i+1}$ to form its contextual representation $h_i$:
\begin{equation}
    h_{i}=\left[f_{i};~ b_{i+1}\right].
\end{equation}

In biaffine model, the role of each word could be either the start of a span or the end of a span. After yielding the contextual representation $h$ of each word, we implement four MLP (multi-layer perceptron) modules for each word, resulting in four vectors, $v_{span_l}$, $v_{span_r}$, $v_{label_l}$ and $v_{label_r}$ (collectively referred to as  $v_{l,i}$, $v_{r,i}$  in Fig.~\ref{fig2}). The first two participate the span-parsing  model deciding whether a span exists or not while $v_{label_l}$ and $v_{label_r}$ are used to label the span with POS tag. The dimension size of $v_{span_l}$ and $v_{span_r}$ is 450, while that of $v_{label_l}$ and $v_{label_r}$ is 100.

\subsection{Span parsing.}
We construct an entity-aware biaffine attention model for constituent parsing. In our baseline biaffine model \cite{zhang2020fast}, $word_i$ always uses the same start vector $v_{span_{l,i}}$ when pairs with different end $word_j$ or $word_k$. We design a unique start and end vector for each $span_{(i,j)}$ based on start vector of $word_i$ i.e. $v_{span_{l,i}}$, end vector of $word_j$ i.e. $v_{span_{r,j}}$ and entity information of $span_{(i,j)}$. More specifically, for a given sentence with $n$ words, we construct two $n*n$ embedding matrixes, embed-matrix-l: $L$ and embed-matrix-r: $R$, which store the start and end vectors for all possible spans. For example, $L[i,j]$ and $R[i,j]$ are the start and end vectors applied in biaffine calculation of $span_{(i,j)}$. As a consequence, the issue left now is how to build $L$ and $R$? For a given sentence with $n$ words, the $n*n$ matrix $S$ represent all possible spans. $S[i,j]$ means a span start from $word_i$ and end at $word_j$. It is obvious that these nodes in the Lower triangle of matrix $S$ bear no meaning (we cannot build a span of words starts from the back and ends at the front), which will not be considered in practice. We traverse every node located in the upper triangle of $S$. For a certain node $S[i,j]$, we check whether $span_{(i,j)}$ is an entity, and embed entity information (i.e. 0 or 1) into a vector $v_{is\_entity}$ with length 50. Then we concatenate $v_{is\_entity}$ to $v_{span_{l,i}}$ and $v_{span_{r,j}}$, resulting in two vector $v_{l(i,j)}$ and $v_{r(i,j)}$ of length 500. Store $v_{l(i,j)}$ in $L$ and $v_{r(i,j)}$ in $R$ at position $(i,j)$, respectively.  
When doing biaffine operation, we traverse the upper triangle of matrix $S$ again, for
every $node(i,j)$, and fetch the start vector $v_{l(i,j)}$ from $L$ and end vector $v_{r(i,j)}$ from $R$ at position $(i,j)$. Then both $v_{l(i,j)}$and $v_{r(i,j)}$  are fed to the following biaffine operation:
\begin{equation}
    v_{l(i, j)}^{T} W v_{r(i, j)},
\end{equation}
where $v_{l(i,j)}$ and $v_{r(i,j)}$ are vectors with length $d$=500, and $W$ is the learning parameter with $d*d$. The result is a scalar indicating the potential score for $span_{(i,j)}$ being a node in the sentence’s constituent parsing tree, and we mark it as $s(i,j)$.

For a sentence $\mathbf{x}$, the score of a tree $\mathbf{y}$ is the sum score of all spans containing in the tree:

\begin{equation}
\label{eq4}
s(\mathbf{x}, \mathbf{y})=\sum_{(i, j) \in \mathbf{y}}s(i,j).
\end{equation}

Under TreeCRF algorithm, the condition probability of the golden tree is:
\begin{equation}
    p(\mathbf{\hat{y}} \mid \mathbf{x})=\frac{e^{s(\mathbf{x}, \mathbf{\hat{y} })}}{Z(\mathbf{x}) \equiv \sum_{\mathbf{y}^{\prime} \in \mathcal{T}(\mathbf{x})} e^{s\left(\mathbf{x}, \mathbf{y}^{\prime}\right)}},
\end{equation}
where $Z_{(x)}$ is the sum score for all possible trees, which can be calculated by inside algorithm. $\mathcal{T}(\mathbf{x})$ is a set including all possible trees for $\mathbf{x}$.

Based on the scores of all spans calculated above, we use the CKY algorithm to decode the parse tree with the highest score:

\begin{equation}
    \mathbf{\bar{y}}=\arg \max _{\mathbf{y}} s(\mathbf{x}, \mathbf{y})=\arg \max _{\mathbf{y}} p(\mathbf{y} \mid \mathbf{x}).
\end{equation}

As illustrated above, we build unique start and end vectors for each span by adding their entity attribute embedding(whether it is an entity or not). Through this processing, parsing inference can then consider the entity information, which is entity-aware.

\subsubsection{Label parsing.}
After finishing the span parsing work, we obtain a constituent tree structure without labels. We feed $v_{label_{l,i}}$ and $v_{label_{r,j}}$ into the following biaffine attention operation to finish the labeling job for $span_{(i,j)}$:

\begin{equation}
    v_{label_{l,i}}^{T} W v_{label_{r,j}}.
\end{equation}

In this module, the corresponding parameter $W$ is $c*d*d$, where $c$ is the number of POS types and $d$ is the length of  $v_{label_{l,i}}$ and $v_{label_{r,j}}$. The result is a probability vector of length $c$. 

\subsection{Entity compatible split method}
Our model belongs to chart-based parsing model. It relies on CKY algorithm to decode. The result trees are binarized trees agreeing with CNF rule. However, little training data satisfies a binarized tree structure. It is common to use third-party tool (e.g. NLTK) to convert an original parsing tree into its binarized form with split choice right or left. As we observed, many entities satisfy a sub-tree structure in their original non-binarized parsing trees, violating after these trees binarized. One group of samples suffers from left binarized operation, while other samples suffering from right choice. The baseline biaffine attention model \cite{zhang2020fast} uniformly chooses left split mechanism, which is not friendly to first group. In our model, we compare entity violating number of the two split choices for every single sample, and choose those results suffering from lower entity violating number.

\subsection{NER sub-task}
Besides adding entity information in parsing process, to make our model more related with entity structure, we add a bi-nary NER model that only judges whether a span is an entity or not. The NER model is also implemented with biaffine attention architecture, which shares the same contextual word embedding $h$ with parsing model. Computing entity score of a certain $span_{(i,j)}$ is analogous to parsing model above. Two extra MLPs are applied to obtain entity span boundaries representation $v_{entity_l}$ and $v_{entity_r}$, then follows the same biaffine process as Label parsing, $v_{entity_{l,i}}^TWv_{entity_{r,j}}$. $W$ is a $2*d^{\prime}*d^{\prime}$ tensor, 2 represents two types (a span is an entity or not). $d^{\prime}$ is the dimensions of $v_{entity_{l,i}}$, $v_{entity_{r,j}}$ and is set to 150 in this paper. 

\subsection{Training loss}
The losses for the whole model originate from three parts: Span parsing loss, Label parsing loss and NER loss:

\begin{equation}
 Loss=Loss_{span}+Loss_{label}+Loss_{entity}.
\end{equation}

The first item $Loss_{span}$ is to maximize the probability of the golden parsing tree and has the following form:
\begin{equation}
L_{span}(\mathbf{x}, \mathbf{\hat{y}})=-s(\mathbf{x}, \mathbf{\hat{y}})+\log Z(\mathbf{x}).
\end{equation}

The latter two terms correspond to label parsing loss and NER loss, which belong to classification tasks and have the common cross entropy loss, i.e.,
\begin{equation}
p(i, j)_{c}=\frac{\exp \left(score(i, j)_{c}\right)}{\sum_{c^{\prime}=1}^{C} \exp \left(score(i, j)_{c^{\prime}}\right)},
\end{equation}

\begin{equation}
loss=-\sum_{(i,j)\in {\hat{y}}} \sum_{c^{\prime}=1}^{C} y_{(i, j)_{c^{\prime}}} \log p_{(i, j)_{c^{\prime}}},
\end{equation}
where $p(i,j)_{c}$ is the probability of entity $span_{(i,j)}$ with label $c$ in NER task, or the parse span probability with label $c$ for label parsing task. $y(i,j)$ is a one-hot vector with golden label position setting 1 and the other positions setting 0. $\hat{y}$ here can be a golden parsing tree or a set containing all golden entity spans of sentence $x$.

\section{Experiment}
\subsection{Dataset.}
We conduct experiments on PTB, CTB, and ONTONOTES. The first two datasets are popular benchmarks in constituent parsing tasks, while ONTONOTES contains both parsing and NER tags, which is rather suitable for discussing the entity-violating issue. Given that there is no NER tags in datasets PTB and CTB5.1, we use third party tool StanfordCoreNLP to obtain the entities on these two datasets. We follow the conventional train/dev/test data split approaches on the three datasets. 

\subsection{Metrics.}
The main idea of this paper is to alleviate the entity-violating issue in constituent parsing task. Besides the following frequently-used three metrics, i.e. precision, recall and F1-score, we introduce a new metric named entity-violating rate named $EVR$, which indicates how many samples suffer from entity violating problem. We calculate $EVR$ as follows:

\begin{equation}
EVR=num_{v}/num_{s},
\end{equation}
where $num_{v}$ is the number of entities conflicting with constituent trees, and $num_{s}$ is the total number of samples.

\subsection{Parameter Setting.} 
To compare with baseline biaffine attention method and illustrate the effectiveness of our model in $EVR$ aspect, we follow most of hyper-parameter values in \cite{zhang2020fast}. The main parameters are shown in Table \ref{tab1}. We set all the dropout rate to 0.33, and the batch size to 1000, respectively. Our model is optimized by Adam, and the learning rate is 0.001 with decay rate 0.999 after every 100 steps. 

\begin{table*}[h]
\centering
 \caption{ Hyper-parameter setting}
 \label{tab1}
 \begin{tabular}{ccccccc}
 \hline
  
  $word_{embed}$ & $feature_{embed}$ & $BiLSTM_{hidden}$ & $v_{span_{l/r}}$  &  $v_{label_{l/r}}$ & $v_{entity}$ & $v_{entity_{l/r}}$ \\
  \hline
  300 & 100 & 400 & 450  &  100 & 50 & 150\\
  
  \hline
 \end{tabular}
\end{table*}

\subsection{Compared Models.}
We compare our model with 5 methods that have ranked among the best in constituent parsing. The abbreviation of model names are shown in Table \ref{tab2}.  B-biaffine{\cite{zhang2020fast}} is the abbreviation of baseline biaffine model, it achieves constituent parsing through basic biaffine attention. Benepar{T5\cite{kitaev2018constituency}} is proposed by Kitaev and Klein, it encodes spans with self-attention, using MLP to obtain the confidence score of a span to be a node in constituent tree. LAL{XLNet\cite{mrini2019rethinking}} is a model that represents spans with label attention and follows the same decoding framework as Benepar{T5\cite{kitaev2018constituency}}. SAPar{Bert\cite{tian2020improving}} is also a chart-based model, it encodes spans by n-gram attention. HPSG{Bert\cite{zhou2019head}} adjusts the form of Head-Driven phrase structure grammar (HPSG) to satisfy both constituent parsing and dependency parsing, and fulfills a joint constituent and dependency parsing model sharing syntactic information of each task.

Given that Benepar{T5\cite{kitaev2018constituency}}, LAL{XLNet\cite{mrini2019rethinking}}, SAPar{Bert\cite{tian2020improving}} and HPSG{Bert\cite{zhou2019head}} have not been trained on ONTONOTES dataset, we just run prediction on PTB and CTB for these four models. For Benepar{T5\cite{kitaev2018constituency}}, we run parsing operation based on the published benepar tool package. For LAL{XLNet\cite{mrini2019rethinking}}, SAPar{Bert\cite{tian2020improving}} and HPSG{Bert\cite{zhou2019head}}, we download their published pre-trained models on websites corresponding to PTB and CTB datasets.


\begin{table*}[h]
\centering
 \caption{Results on ONTONOTES and PTB}
 \label{tab2}
 \begin{tabular}{lcccccccc}
 \hline
  
  ~ & \multicolumn{4}{c}{ONTONOTES} & \multicolumn{4}{c}{PTB}   \\
  
  & P & R & F1 & \multicolumn{1}{c}{$EVR$}  &P & R & F1 & $EVR$\\
 \hline
  Benepar{T5\cite{kitaev2018constituency}} & - & - & - & - & 93.22 & 93.06 & 93.14 & 18.53 \\
  SAPar{Bert\cite{tian2020improving}} & - & - & - & - & 95.64 & 95.64 & 95.64 & 17.73 \\
  LAL{XLNet\cite{mrini2019rethinking}} & - & - & - & - & 94.43 & 94.34 & 94.39 & 18.70 \\
  HPSG{Bert\cite{zhou2019head}} & - & - & - & - & \textbf{95.69} & \textbf{95.69} & \textbf{95.69} & 17.57  \\
  B-biaffine{\cite{zhang2020fast}} & 91.44 & 91.42 & 91.43 & 2.64 & 93.79 & 93.85 & 93.82 & 17.60  \\
  \hline
  Ours{GC} & 92.15 & 92.32 & 92.23 & {[}\textbf{0.65}{]} & 93.54& 93.91 & 93.72 & 12.51  \\
 $Ours{GB}$ & 95.36 & 95.02 & 95.18 & 1.10 & 94.91 & 94.54 & 94.72 & {[}\textbf{10.29}{]} \\
  $Ours{B}$ & \textbf{95.34} & \textbf{95.49} & \textbf{95.41} & 0.98 & 94.55 & 94.45 & 94.50 & 12.12 \\
 $Ours{right}$ & 92.13 & 92.40 & 92.26 & 1.10 & 93.95 & 93.84 & 93.89 & 17.58  \\
 $Ours{noNER}$ & 92.03 & 92.19 & 92.11 & 1.21 & 93.36 & 93.45 & 93.40 & 12.96 \\
  
  \hline
 \end{tabular}
\end{table*}


\begin{table*}[h]
\centering
 \caption{Results on CTB5.1}
 \label{tab3}
 \begin{tabular}{lcccc}
 \hline
  
  ~ & \multicolumn{4}{c}{CTB} \\
  
  & P & R & F1 & \multicolumn{1}{c}{$EVR$} \\
 \hline
 
  SAPar{Bert\cite{tian2020improving}} &\textbf{92.62} & \textbf{92.62} & \textbf{92.62} & 16.51\\
  B-biaffine{\cite{zhang2020fast}} &88.53 & 88.64 & 88.58 & 17.14\\
  \hline
  $Ours{B}$ & 88.42 & 89.71 & 89.06 & {[}\textbf{14.92}{]}\\
 $Ours{right}$ &88.30 & 88.58 & 88.44 & 16.50\\
 $Ours{noNER}$ &88.75 & 88.75 & 88.75 & 15.87\\
  
  \hline
 \end{tabular}
\end{table*}


\subsection{Results.}
The results of three conventional metrics precision, recall, F1-score and our introduced $EVR$ are shown in Table \ref{tab2} and Table \ref{tab3}. In each experiment performance was averaged over seven runs. The superscripts of these compared models indicate the pre-trained embeddings, while in the lower part of our models, these superscripts bear the following meaning:
$Ours{GC}$ means that we use glove word embedding concatenating char-level features as initial input, which is also our kernel method mentioned above. $Ours{GB}$ replaces char-level feature with Bert-feature, and $Ours{B}$ initializes word embedding randomly (without glove), using Bert embedding as feature. $Ours{right}$ is a variant based on $Ours{GC}$, using right-choice binarized tree to compare with our proposed entity compatible split method. $Ours{noNER}$ cuts off the NER model of $Ours{GC}$.

\indent \textbf{Comparison with other methods:}
When comparing our proposed model with the other five baseline models, we can see that our proposed models outperform all the other models on metric $EVR$. The observations are detailed as follows: a) On ONTONOTES dataset, $Ours{GC}$ reduces more than 75\% of $EVR$ when compared with $B-biaffine$, while maintaining a higher level of the other three metric(precision, recall ,F1-score). b) On PTB dataset, $HPSG{Bert}$ obtains the highest performance on F1-score with the pre-trained Bert embeddings and shared syntactic information between constituent and dependency parsing. However, these 5 models perform worse more than 5 points of $EVR$ than our proposed models $Ours{GC}$, $Ours{GB}$. c) On CTB dataset, $Ours{B}$ achieves lower F1-score than $SAPar{Bert}$, may be due to the inappropriate way of Bert feature used in our model, however we still gain the best performance on $EVR$ indicator. d) The $EVR$ on ONTONOTES is much lower than that on PTB and CTB. Since ONTONOTES is a professional NER dataset with high quality NER labeled data while we apply the third-party tool on PTB and CTB to get NER labels, which has an unexpected result with noise.

\indent \textbf{Ablation study:}
a) When comparing $Ours{GC}$ with $Ours{GB}$ across the three datasets, the latter improves performance on PTB and CTB datasets after using Bert features while $Ours{GC}$ achieves the lowest violating rate on ONTONOTES. It indicates that our defined $EVR$ is not that sensitive to these pre-trained embeddings. b) When the original binarized tree method is applied, $Ours{right}$ suffers from higher $EVR$ than $Ours{GC}$ across all the three datasets, which proves the effectiveness of our proposed compatible split method. c) $Ours{noNER}$ cuts off the NER sub-task module and performs worse than $Ours{GC}$, which suggests that the NER sub-task helps our model to understand the added NER feature better.

\begin{table}
\centering
\caption{Comparison of downstream task performance }
\label{tab4}
\begin{tabular}{cc}
\hline
Parsing Data & Accuracy \\
\hline
SAPar{[Bert\cite{tian2020improving}]} &  95.4 \\
HPSG{[Bert\cite{zhou2019head}]} &  96.0 \\
LAL{[XLNet\cite{mrini2019rethinking}]} & 96.2 \\
Benepar{[T5\cite{kitaev2018constituency}]} & 95.4 \\
B-biaffine{[\cite{zhang2020fast}]} & 95.0  \\
Ours & \textbf{96.2}  \\
\hline
\end{tabular}
\end{table}

\subsection{\textbf{Performance in downstream tasks.}}
To make our method and the introduced $EVR$ metric more convincing, we extend our parsing model to a downstream task: Sentiment Analysis.
Kim et al. \cite{kim2019dynamic} introduced a Tree-LSTM framework for sentence sentiment classification based on constituent parsing tree. It implements bottom-up LSTM operation recursively and sends the root node embedding into a inference layer for classification results.
We deploy the tree structure used in \cite{kim2019dynamic} with the counterparts from Benepar{T5\cite{kitaev2018constituency}}, LAL{XLNet\cite{mrini2019rethinking}}, SAPar{Bert\cite{tian2020improving}}, HPSG{Bert\cite{zhou2019head}} and B-biaffine{\cite{zhang2020fast}}, and our proposed model, respectively, in order to compare the effectiveness of parsing tree generated by our model.

Tabel \ref{tab4} depicts the results. Our model and LAL{XLNet\cite{mrini2019rethinking}} achieve the highest accuracy 96.2\% on sentence classification on TREC. It implies that our entity-aware biaffine attention model is more in line with the language model. 

\subsection{Case study.}
\begin{figure}
\setlength{\abovecaptionskip}{-0.2cm}
\centering
\includegraphics[width=1.0\textwidth]{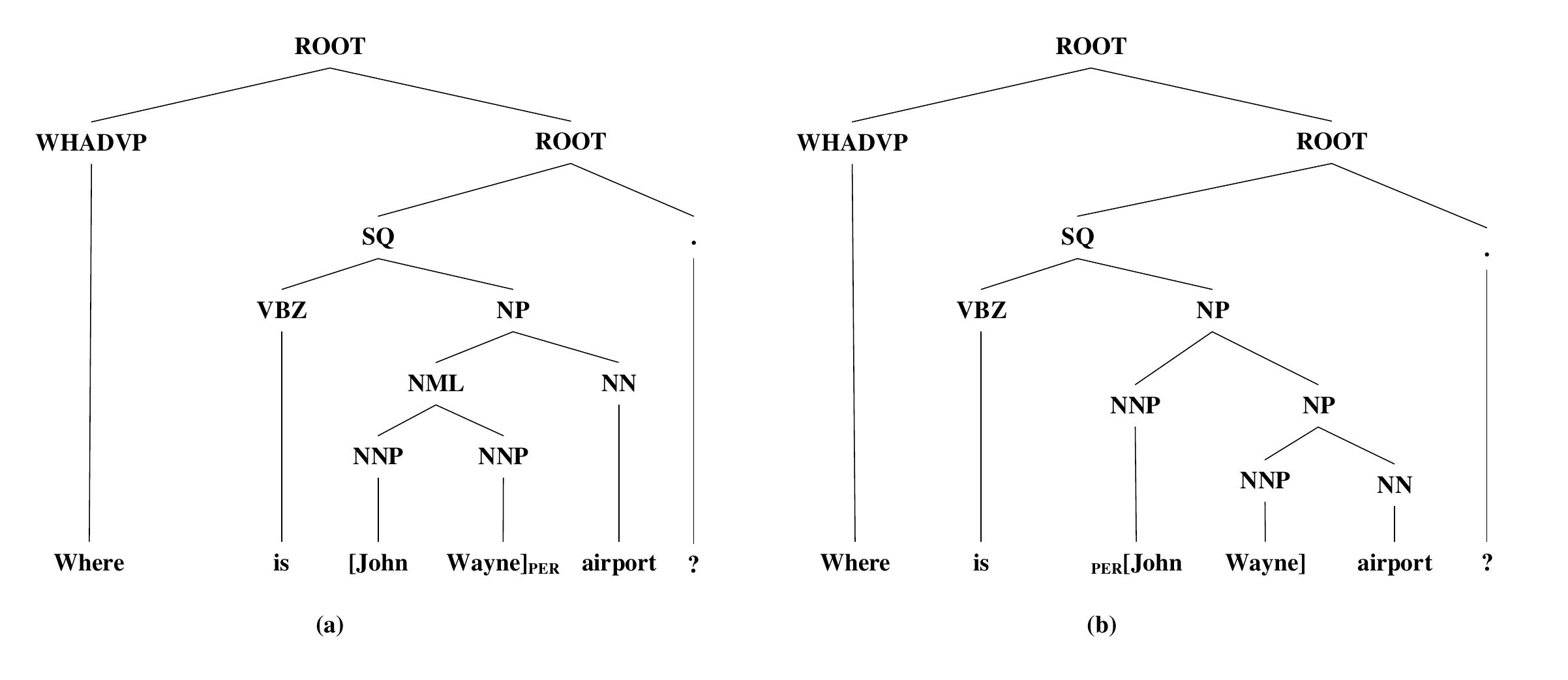}
\caption{Comparison of the parsing result for "Where is John Wayne airport?" between (a) our model and (b) B-biaffine{\cite{zhang2020fast}}. 
\label{fig3}}
\end{figure}

Figure~\ref{fig3} illustrates a case study performed by our proposed model (Fig.~\ref{fig3} (a)) and the baseline biaffine attention model (Fig.~\ref{fig3} (b)) for sentence “Where is John Wayne airport?”. Our model treats the PERSON entity “John Wayne” as a complete constituent, while the baseline model splits it, and integrates the two words "Wayne" and "airport" into a sub-tree. The example indicates that our model can learn additional entity structure knowledge. 

\section{Conclusion}
In this paper, we investigate entity-violating problem in constituent parsing tasks. To alleviate the violating issue, we construct an entity-aware parsing model based on biaffine attention method. We modify the basic biaffine model, making every biaffine operation correlated with its span’s entity information, without extra manual annotations. Experimental results on ONTONOTES, PTB and CTB show that our proposed model achieves lowest $EVR$ on the three datasets. The best performance of our parsing model on downstream task also demonstrates the superiority of our method.

\bibliography{iclr2024_conference}
\bibliographystyle{iclr2024_conference}

\end{document}